# CURRENT: A Recurrent Neural Network for Concurrent Spatial and Temporal Super-Resolution of 4D Flow MRI

Luuk Jacobs[1*], Pietro Dirix[1], Sebastian Kozerke[1]
[1] Institute for Biomedical Engineering, University and ETH Zurich, Zurich, Switzerland
[*] Corresponding author

## 1 Abstract

**Objective:** To improve the spatial and temporal resolution of aortic 4D flow MRI through a unified deep learning framework for post-acquisition super-resolution (SR) of reconstructed velocity vector fields.
**Methods:** We propose CURRENT, a recurrent neural network that performs both spatial and temporal super-resolution with a single flexible network architecture. Supervised training was conducted using paired low- and high-resolution datasets generated by embedding computational fluid dynamics simulations of aortic flow into in vivo thoracic MRI backgrounds, followed by controlled spatial and temporal downsampling. CURRENT was evaluated for two-fold spatial and temporal SR against 4DFlowNet and for two-fold spatiotemporal SR against sequential spatial and temporal SR pipelines based on 4DFlowNet. Image-based and k-space-based quantitative metrics were considered and conventional interpolation were used as baselines. In addition, the impact of CURRENT's spatiotemporal SR on physics-informed relative pressure estimation was assessed. Finally, the ability of CURRENT to generalize from synthetic training data to in vivo acquisitions was investigated.
**Results:** CURRENT achieved superior spatial SR performance and comparable temporal SR performance to dedicated 4DFlowNet models while providing a unified solution for both tasks. Simultaneous spatiotemporal SR using CURRENT yielded more accurate relative pressure estimates than either spatial or temporal SR alone and outperformed sequential 4DFlowNet-based spatial and temporal SR approaches. Furthermore, the proposed model demonstrated promising generalization to in vivo 4D flow MRI data despite being trained exclusively on synthetic datasets.
**Conclusion:** CURRENT provides a unified framework for spatial and temporal SR of 4D flow MRI. The proposed approach has the potential to enhance physics-based analysis and broaden the applicability of accelerated 4D flow MRI.

## 2 Keywords

# 3 Introduction

4D flow MRI [1] is an established technique to quantify 3D velocity vector fields resolved along the cardiac cycle. The accuracy and precision of velocity quantification depends on the acquired spatiotemporal resolution and signal-to-noise ratio (SNR) [2], which in turn relate to acquisition times. To improve resolution for a given acquisition time, deep learning (DL) has been deployed to perform post-acquisition super-resolution (SR) and denoising. Generally, DL-based SR methods can be separated into physics-informed neural networks (PINNs) and convolutional neural networks (CNNs). While PINNs perform a time-consuming iterative approximation of the Navier-Stokes equations per anatomy to predict high-frequency information, CNNs offer clinically acceptable inference times by predicting high-frequency information from statistical priors learned from redundancies and structural correlations present in the training data. Accordingly, CNNs can generalize to diverse geometries without the need for patient-specific retraining. Typically, these CNNs are trained and validated on synthetic data, given that in vivo data at the desired high resolution (HR) are infeasible to acquire. More specifically, geometries and boundary conditions are either generated or extracted from in vivo 4D flow MRI data and are used to simulate HR velocity fields using computational fluid dynamics (CFD). These synthetic datasets are subsequently downsampled to generate paired low resolution (LR) data [3].

For spatial SR of 4D flow MRI (Fig. 1a), 4DFlowNet, a CNN-based approach has been proposed (hereafter referred to as S4DFlowNet to note its task of spatial SR) [3]. Its architecture consists of multiple residual layers using 3D spatial convolutions with a fixed upsampling layer in between. S4FlowNet has been deployed for various applications of spatial SR of 4D flow MRI [4], [5], [6], [7] and demonstrates comparable performance to band-limited (sinc) interpolation for systolic frames and improved performance for diastolic frames for an unseen in silico case. However, its frame-wise approach may limit performance as temporal correlations present in the 4D flow MRI data are not exploited. So far, only two CNN-based spatial SR approach exist that are not strictly frame-wise: 1) SURFR-Net [8], which implicitly injects some temporal information by using the Weighted Mean Frequencies (WMF) [9] as an additional input feature channel to the three velocity components per cardiac bin and 2) tempo-mDCSRN-Flow/TransVort [10], which implements a long short-term memory (LSTM) [11] network architecture for improved temporal coherence for spatial SR of cerebrospinal fluid (CSF) flow in static cerebral ventricles.

For temporal SR of 4D flow MRI (Fig. 1b), Callmer et al. [12] have proposed a CNN-based approach by modifying the S4DFlowNet architecture to operate on 2D+time patches instead of 3D patches, thereby upsampling the time dimension (hereafter referred to as T4DFlowNet to note its task of temporal SR). It was shown that T4DFlowNet outperformed linear and sinc interpolation in terms of mean absolute error (MAE) for an unseen in silico case. However, the omission of one of the three spatial dimensions for each forward pass may limit its performance.

Spatiotemporal SR of 4D flow MRI (Fig. 1c) using a CNN has not been implemented so far, despite its relevance for e.g. non-invasive estimation of relative pressure to assess aortic stenosis [13], coarctation [14], and dissection [15]. Since relative pressure estimation relies on both spatial and temporal velocity derivatives, high spatiotemporal resolution is required. Particularly for the aorta, where the higher pulsatility results in higher temporal velocity gradient compared to more distal vasculatures, a high temporal resolution is critical. Of note, in the non-medical SR literature [16], two-stage approaches have been considered, i.e. decomposing the problem into two independent sequential tasks of temporal and spatial SR, which may, however, not fully exploit intrinsic correlations between the two tasks.

In this work, we propose a recurrent neural network (CURRENT) that can perform spatial, temporal, and spatiotemporal SR of 4D flow MRI by operating on 4D (3D+time) patches of velocity vector fields, thereby incorporating I) spatial SR that explicitly exploits the sequential cardiac dimension, II) temporal SR that exploits all three spatial dimensions, and III) CNN-based joint spatiotemporal SR.

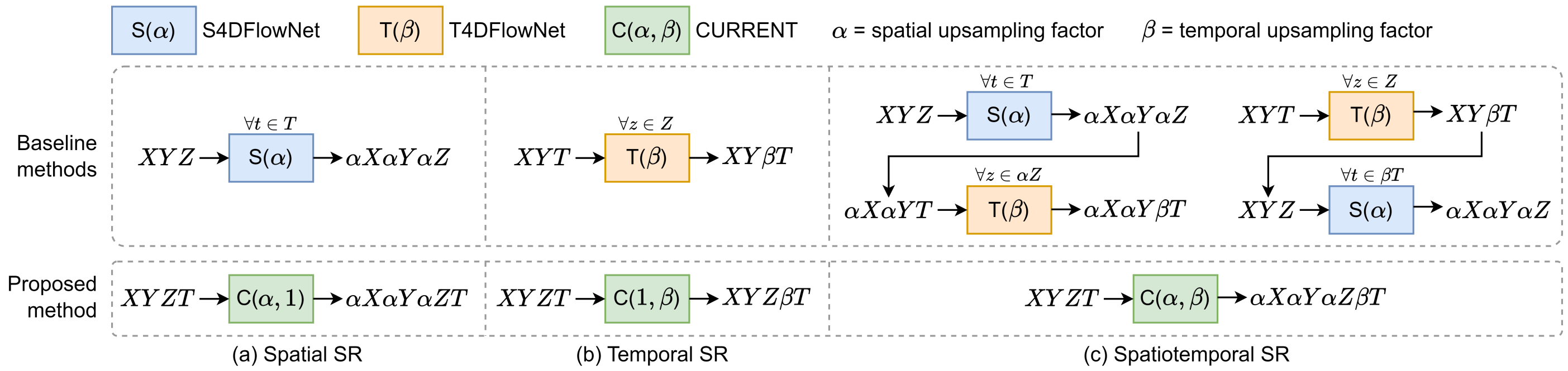

***Figure 1: Overview of super-resolution (SR) methods.*** *The three spatial dimensions and cardiac dimension are represented by $X$, $Y$, $Z$, and $T$, respectively. (a) Spatial SR is performed by S4DFlowNet in a frame-wise manner. (b) Temporal SR is performed by T4DFlowNet by omitting one of the spatial dimensions; temporal SR of $XZT$ and $YZT$ are omitted for visualization purposes. (c) Spatiotemporal SR can be performed by using S4DFlowNet and T4DFlowNet in a two-stage approach. Note that the same CURRENT framework can be used for spatial (α>1, β=1), temporal (α=1, β>1), and spatiotemporal (α>1, β>1) SR.*

# 4 Methods

## 4.1 Data processing

For each cardiac frame, 10 patches of $N_P \times N_P \times N_P$ voxels were randomly extracted from the LR data, with 9 out of 10 patches containing a minimum of 20% fluid-containing voxels. Corresponding patches of $2N_P \times 2N_P \times 2N_P$ voxels were then extracted from the HR data at the same spatial location to create aligned patch pairs. A sliding window approach was used along the cardiac dimension with window size of $N_T$ cardiac bins and the three velocity components were concatenated in the input feature channel dimension and normalized by their velocity encoding (VENC). Each patch had 10 augmented versions generated by rotation of 90°, 180°, and 270° in all three spatial directions, resulting in 2.5K samples per geometry. The data was split into training (N=10, 25K samples), validation (N=5, 2.5K samples), and testing (N=10) sets.

Inference was performed by extracting patches from LR data with a stride of $(N_P - 4)$ in all three spatial dimensions, retaining only the central $(N_P - 4)$ voxels per dimension, and temporal windows with a stride of $(N_T - 4)$, retaining only the central $(N_T - 4)$ cardiac bin(s), stitching together the final super-resolved output. Spatiotemporal edges are handled accordingly to ensure a complete two-fold upsampling.

## 4.2 Network architecture

An overview of CURRENT's architecture is presented in Fig. 2a, taking noisy LR data $\boldsymbol{v_{LR}} \in \mathbb{R}^{N_p \times N_p \times N_p \times N_t \times 3}$ as input (see 4.1 Data processing) with the three velocity components in the feature dimension and outputting denoised and spatially and/or temporally super-resolved data $\boldsymbol{v_{SR}} \in \mathbb{R}^{\alpha N_p \times \alpha N_p \times \alpha N_p \times \beta N_t \times 3}$ through a sequence of 3D spatial convolutions surrounding a central upsampling layer parameterized by spatial upsampling factor $\alpha$ and temporal upsampling factor $\beta$. CURRENT has a single input path (shared-encoding) for computational efficiency and three output paths (separated decoding) to predict each velocity component [17]. The main architectural modifications compared to S4DFlowNet and T4DFlowNet include: I) bidirectional convolution recurrent neural network (BCRNN) [18] layers instead of residual blocks to fully exploit all correlations available in the 3D+time velocity vector fields, II) replacing the standard skip connection ($y = x + layer_n(x)$) with a learnable skip connection ($y = x + \rho_n layer_n(x)$, with learnable scaling factors $\rho_n$), and III) adding long learnable skip connections (with learnable scaling factors $\rho_{LR}$ and $\rho_{HR}$) for enhanced gradient flow via residual learning on a coarse level [19]. The optimized scaling factors for CURRENT are presented in Fig. 2b, showing distinct patterns for two-fold spatial, temporal, and spatiotemporal SR.

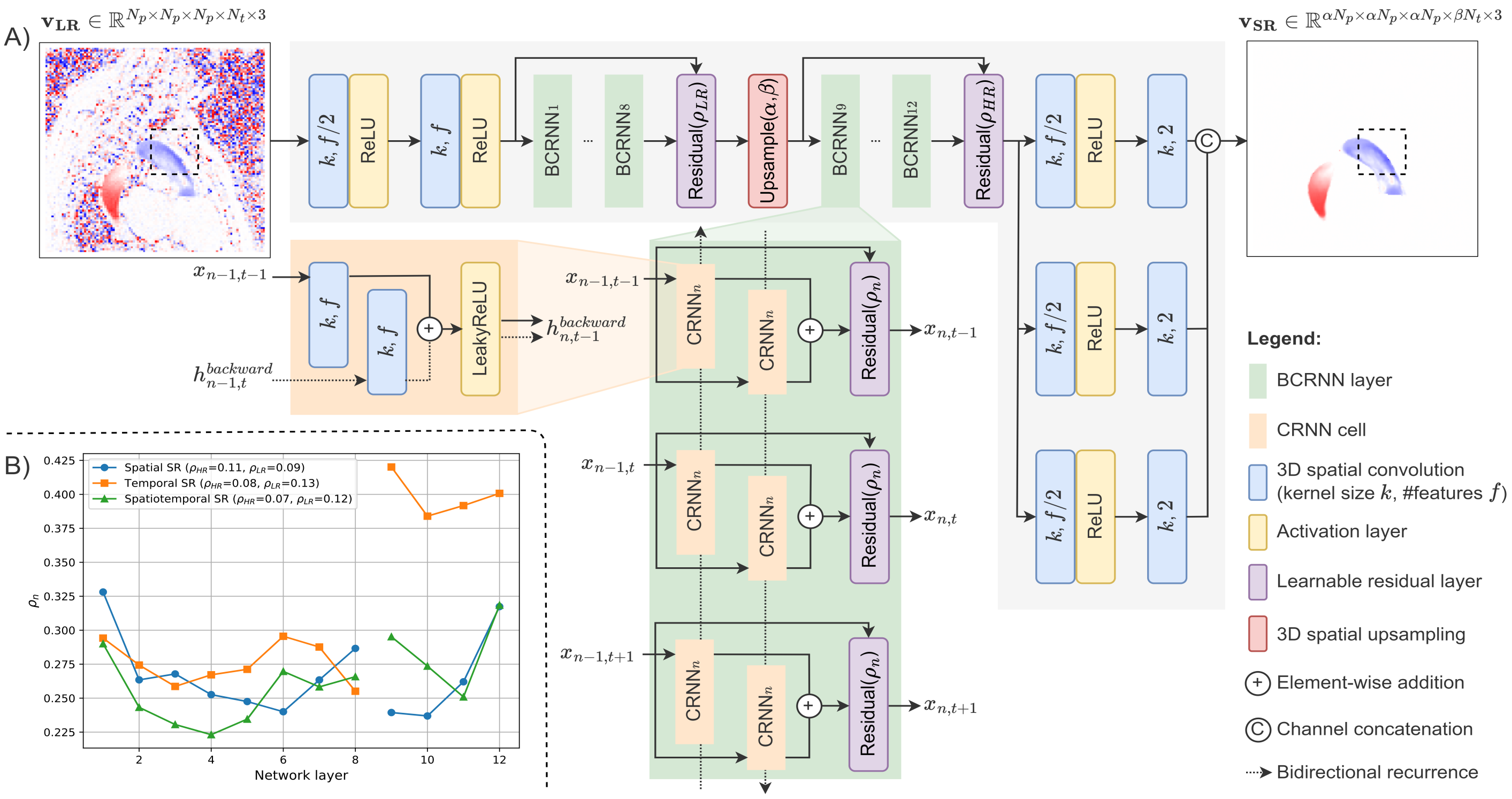

***Figure 2: Overview of CURRENT architecture.*** *A) The network receives a temporal window of cubic patches (dashed black box) as input and incorporates BCRNN layers (passing hidden states between preceding [backward pass] and subsequent [forward pass] cardiac bins) and learnable short/long skip connections surrounding a central bicubic upsampling. B) The optimized scaling factors for the short ($\rho_n$ for the $n^{th}$ BCRNN layer) and long ($\rho_{LR}$ and $\rho_{HR}$) skip connections. Note that the same CURRENT framework can be used for spatial (α>1, β=1), temporal (α=1, β>1), and spatiotemporal (α>1, β>1) SR.*

### 4.3 Training loss

A voxel-wise mean-square error (MSE) loss $\mathcal{L}_{MSE}(\boldsymbol{v_{SR}}, \boldsymbol{v_{HR}})$ was used to compare $\boldsymbol{v_{SR}}$ to ground truth $\boldsymbol{v_{HR}} \in \mathbb{R}^{2N_p \times 2N_p \times 2N_p \times N_t \times 3}$ in regions $\Omega \in \{\Omega_{non-fluid}, \Omega_{fluid}\}$:

$$\mathcal{L}_{MSE}(\boldsymbol{u}, \boldsymbol{v}) = \frac{1}{N_\Omega} \sum_{i \in \Omega} \left(\boldsymbol{u}^{\boldsymbol{i}} - \boldsymbol{v}^{\boldsymbol{i}}\right)^2, \tag{1}$$

where $N_\Omega$ is the number of voxels in region $\Omega$. Additionally, a voxel-wise mutually projected loss $\mathcal{L}_{mp}(\boldsymbol{v_{SR}}, \boldsymbol{v_{HR}})$ was included for the fluid regions to increase directional consistency [20]:

$$\mathcal{L}_{mp}(\boldsymbol{u}, \boldsymbol{v}) = \frac{1}{N_{\Omega_{fluid}}} \sum_{i \in \Omega_{fluid}} \frac{1}{2}\left(\left|\|\boldsymbol{u}^{\boldsymbol{i}}\| - \|\boldsymbol{v}^{\boldsymbol{i}}\| \cos(\theta)\right| + \left|\|\boldsymbol{v}^{\boldsymbol{i}}\| - \|\boldsymbol{u}^{\boldsymbol{i}}\| \cos(\theta)\right|\right), \tag{2}$$

where $\theta$ is the angle between $\boldsymbol{u}^{\boldsymbol{i}}$ and $\boldsymbol{v}^{\boldsymbol{i}}$. Finally, an $l_2$ regularization term on the network weights ($\boldsymbol{w}$) was included to prevent overfitting, resulting in the following total loss used for training all SR networks:

$$\mathcal{L}_{total}(\boldsymbol{u}, \boldsymbol{v}, \boldsymbol{w}) = \lambda_{MSE}\mathcal{L}_{MSE}(\boldsymbol{u}, \boldsymbol{v}) \; + \; \lambda_{mp}\mathcal{L}_{mp}(\boldsymbol{u}, \boldsymbol{v}) \; + \; \lambda_{reg}\|\boldsymbol{w}\|_2^2, \tag{3}$$

where weights $\lambda_{MSE}$, $\lambda_{mp}$, and $\lambda_{reg}$ determine the relative weights of each loss term. Preliminary experiments showed that a gradient-based divergence loss (based on first order differences) lead to an underestimation of near-wall velocities, as also reported in related works [4], [5], and was hence omitted.

### 4.4 Data synthesis

Synthetic thoracic aorta 4D flow MRI datasets were generated as a subset of the RACLETTE dataset [21], [22], providing paired HR/LR complex-valued in vivo images embedding CFD-generated ground truth aortic flow fields. Data synthesis involved following steps:

1. **Simulating flow fields:** Aortic geometries (N=25) were drawn from publicly available anatomies [23], [24]. Pulsatile inlet boundary conditions were generated from a principal component analysis-based statistical model of in vivo 2D flow MRI waveforms and imposed through a parametrised stenotic aortic-valve model [22], yielding an equal number of healthy, mild, moderate, severe, and critical aortic stenosis cases [25], [26]. Time-resolved compliant-wall CFD simulations were performed using OpenFOAM [27].

2. **Embedding into in vivo background:** The CFD velocity fields were projected onto a uniform grid with 0.5 mm isotropic voxel size and 10 ms temporal resolution. Phase-difference images were computed from the simulated velocity fields ($\vec{v}$), with encoding directions ($\vec{n}$) and VENCs chosen to produce a standard 4D flow phase field ($\overrightarrow{\Delta\phi} = \vec{v} \cdot \pi\vec{n}/VENC$). These were combined with the morphed in vivo background phase and assembled with the corresponding magnitude to obtain synthetic 4D flow MRI datasets.
3. **Downsampling:** Temporal downsampling was performed by bin-averaging to 20 ms (HR) and 40 ms (LR). Spatial downsampling was performed by rectangular boxcar cropping in k-space to isotropic resolutions of 1.0–1.25 mm (HR) and 2.0–2.5 mm (LR). Complex Gaussian noise (with standard deviation $\sigma$) was added to the LR k-space to reach SNR $= \mu/\sigma = 15$, with signal defined as the mean magnitude inside the aorta. Finally, the three velocity components were extracted from the phase differences of the synthetic data and used as network inputs.

Note that by first projecting to 0.5 mm isotropic resolution and then downsampling twice to generate HR and LR data, partial volume effects are present in both HR and LR, resulting in a better posed SR problem compared to if only LR had partial volume effects. Moreover, the inclusion of realistic MRI backgrounds ensured realistic downsampling effects including signal contamination from surrounding background into the fluid domain via the imaging point-spread function, resulting in small but observable artefacts (see Supplementary material).

### 4.5 Implementation details

CURRENT was implemented using Tensorflow 2.20.0 [28] with a Keras backend and trained with mixed precision (FP16 for computation and FP32 for network parameters) using an Adam optimizer [29], a batch size of 15, and a learning rate of $1 \times 10^{-4}$ with a cosine annealing to $1 \times 10^{-6}$ over 125 epochs (equivalent to 200K iterations). Loss weights were set to $\lambda_{MSE} = 1$, $\lambda_{mp} = 0.5$, and $\lambda_{reg} = 1$ and learnable scaling factors $\rho_n$, $\rho_{LR}$, and $\rho_{HR}$ were initialized at 0.2. The number of learnable parameters was kept the same as original the S4DFlowNet implementation at 3.2M, with kernel size $k = 3$, number of features $f = 68$, eight BCRNN layers in LR space, and four BCRNN layers in HR space. All 3D convolutions were preceded by symmetric padding and had an L2 regularization term to promote sparsity. Patch size $N_P$ was set to 16 and temporal window size $N_T$ was set to 5 due to graphics processing unit (GPU) memory constraints. While two-fold upsampling was investigated in this work, it is noted that the framework can be used for arbitrary upsampling factors, depending on available computational resources. Hyperparameters were optimized based on S4DFlowNet performance and kept constant for all networks (only the batch size was bottlenecked by GPU memory constraints of spatiotemporal SR using CURRENT and thus set to 15 for all networks). Training (including validation in every epoch) of CURRENT took 164, 74, and 298 hours on an NVIDIA A6000 GPU for two-fold spatial, temporal, and spatiotemporal SR, respectively. Inference time depended on the matrix size of $\boldsymbol{v_{LR}}$ but a representative case of ($132 \times 112 \times 32 \times 24 \times 3$) took 4.4, 2.0, and 8.2 minutes for two-fold spatial, temporal, and spatiotemporal SR, respectively.

### 4.6 Experiments

The spatial and temporal SR performance of CURRENT was qualitatively and quantitatively (see 4.7 Evaluation metrics) compared to S4DFlowNet and T4DFlowNet, respectively. Linear and sinc interpolations were used as baseline upsampling approaches. Additionally, CNN-based spatiotemporal SR was investigated using two-stage approaches (S4DFlowNet into T4DFlowNet [2S-ST] and vice versa [2S-TS]) and the single-stage CURRENT. The benefit of spatiotemporal SR was investigated by estimating relative pressure $\Delta P$ between planes in the ascending and descending aorta using the virtual work-energy relative pressure (vWERP) [30] method for spatial SR (with temporal linear interpolation), temporal SR (with spatial linear interpolation), and spatiotemporal SR using CURRENT. The estimated $\Delta P$ values were compared to the $\Delta P$ values derived from the spatiotemporal HR reference via linear regression analyses.

The feasibility of generalizing CURRENT, trained exclusively on synthetic data, to inference on in vivo 4D flow MRI data was investigated. To this end, a healthy volunteer (male, 26 years) was prospectively scanned, upon written informed consent and according to institutional and ethical guidelines, using a Philips 3T system (Philips Healthcare, Best, the Netherlands) with an acquired spatial resolution of 1.25 mm isotropic, an acquired temporal resolution of 24.6 ms, a repetition time (TR) of 4.8 ms, an echo time (TE) of 2.9 ms, and a VENC of 2, 1, and 1 m/s in feet-head, right-left, and anteroposterior directions, respectively. A 3D Cartesian elliptical pseudo-spiral golden angle sampling was used with an effective undersampling factor of 11 (after retrospective binning into three equally populated respiratory states using self-gating, retaining the end-expiratory state for reconstruction). Acquisition time was approximately 60 min. Reconstruction was performed with FlowMRI-Net [31] in a zero-shot manner, leveraging transfer learning by pre-training on publicly available 4D flow MRI data [32], with post-processing including bias field correction using N4ITK [33], concomitant field correction according to [34], eddy current correction using third-order polynomial fitting with M-estimate SAmple Consensus (MSAC) [35], and noise addition in k-space. The reconstructed images were retrospectively downsampled to a spatial resolution of 2.5 mm isotropic and/or a temporal resolution of 49.2 ms and served as input for spatial, temporal, and spatiotemporal SR. SR results were compared to their respective HR component by visual analysis, linear regression, and comparisons of peak velocity curves in the ascending aorta (given the descending aorta of HR in vivo data is less reliable due to inferior SNR). Segmentation of temporally-resolved aortic masks for these analyses were performed using an nnU-Net [36].

### 4.7 Evaluation metrics

Following patch-based SR, the predicted $\boldsymbol{v_{SR}}$ patches were stitched together into a full volume (see 4.1 Data processing) and quantitatively compared to $\boldsymbol{v_{HR}}$ using ten different metrics. The first image-based metrics that were computed include the mean directional error (DE):

$$\mathrm{DE} = \frac{1}{|M|}\sum_{i\in M} 1 - \frac{\left|\boldsymbol{v}_{\boldsymbol{SR}}^{\boldsymbol{i}} \cdot \boldsymbol{v}_{\boldsymbol{HR}}^{\boldsymbol{i}}\right|}{\left\|\boldsymbol{v}_{\boldsymbol{SR}}^{\boldsymbol{i}}\right\|\left\|\boldsymbol{v}_{\boldsymbol{HR}}^{\boldsymbol{i}}\right\| + \varepsilon} \tag{2}$$

and vectorial normalized root-mean-square error ($\mathrm{nRMSE}_v$):

$$\mathrm{nRMSE}_v = \sqrt{\frac{\sum_{i\in M}\left\|\boldsymbol{v}_{\boldsymbol{SR}}^{\boldsymbol{i}} - \boldsymbol{v}_{\boldsymbol{HR}}^{\boldsymbol{i}}\right\|^2}{\sum_{i\in M}\left\|\boldsymbol{v}_{\boldsymbol{HR}}^{\boldsymbol{i}}\right\|^2 + \varepsilon}} \tag{3}$$

of velocities inside the fluid mask ($M$) and for near-wall (NW) velocities, defined as a single-voxel boundary identified via binary erosion ($M_{NW} = M - M_{erode}$), with $\varepsilon = 1e - 10$.

Additionally, peak velocity curves were computed and visualized by extracting the coordinate with the highest velocity magnitude in $\boldsymbol{v_{HR}}$ within an eroded version of $M$ and tracking this coordinate over all cardiac bins for $\boldsymbol{v_{HR}}$ and $\boldsymbol{v_{SR}}$. The relative difference ($\varepsilon_{rel}$) between peak velocity magnitudes was computed as follows:

$$\varepsilon_{rel} = \left|\frac{|\boldsymbol{v_{SR}}|_{max} - |\boldsymbol{v_{HR}}|_{max}}{|\boldsymbol{v_{HR}}|_{max}}\right| \times 100\% \tag{5}$$

Finally, Fourier shell analysis [37] was performed to assess spatial SR performance, computing the coherence ($\gamma$):

$$\gamma(k_{\mathrm{i}}) = \frac{\sum_{k\in k_i}\mathcal{F}\boldsymbol{v}_{\boldsymbol{SR}}^{\boldsymbol{k}}(k)\cdot\mathcal{F}\boldsymbol{v}_{\boldsymbol{HR}}^{\boldsymbol{k}}(k)^*}{\sqrt{\sum_{k\in k_i}|\mathcal{F}\boldsymbol{v}_{\boldsymbol{SR}}^{\boldsymbol{k}}(k)|^2\cdot\sum_{k\in k_i}|\mathcal{F}\boldsymbol{v}_{\boldsymbol{HR}}^{\boldsymbol{k}}(k)|^2 + \varepsilon}} \text{ for } i \in \{1, \dots, N_k\} \tag{5}$$

and energy deviation ($ED$):

$$ED(k_i) = \left|\frac{\sum_{k\in k_i}|\mathcal{F}\boldsymbol{v}_{\boldsymbol{SR}}^{\boldsymbol{k}}(k)|^2}{\sum_{k\in k_i}|\mathcal{F}\boldsymbol{v}_{\boldsymbol{HR}}^{\boldsymbol{k}}(k)|^2 + \varepsilon} - 1\right| \text{ for } i \in \{1, \dots, N_k\} \tag{6}$$

for multiple concentric shells with increasing spatial frequency radii in k-space, with radial wavenumber $k$, $i$-th frequency bin $k_i$, Fourier transform $\mathcal{F}$, complex conjugate $\blacksquare^*$, and $\varepsilon = 1e - 10$. By exclusively considering shells beyond the acquired spatial bandwidth, spatial SR performance metrics ($SR_p$), with $p \in \{\gamma, ED\}$, could be derived. Similarly, temporal SR performance metrics ($SR_p^t$) were computed by comparing 1D concentric "shells" in the temporal frequency domain beyond the acquired temporal bandwidth. Note that noiseless LR data was used as input for the baseline sinc and linear interpolations for these Fourier shell analyses.

# 5 Results

While linear and sinc interpolations resulted in large errors in the background and bulk flow for the synthetic data, the DL-based SR methods effectively denoised the velocity vector fields while largely preserving spatiotemporal dynamics. Minor qualitative differences were noted between the DL-based SR approaches for spatial, temporal, and spatiotemporal SR mainly at the boundaries (Fig. 3). CURRENT showed consistently improved performance compared to S4DFlowNet for spatial SR (Tab. 1) and performed similarly to T4DFlowNet and two-stage approaches for temporal (Tab. 2) and spatiotemporal (Tab. 3) SR, respectively.

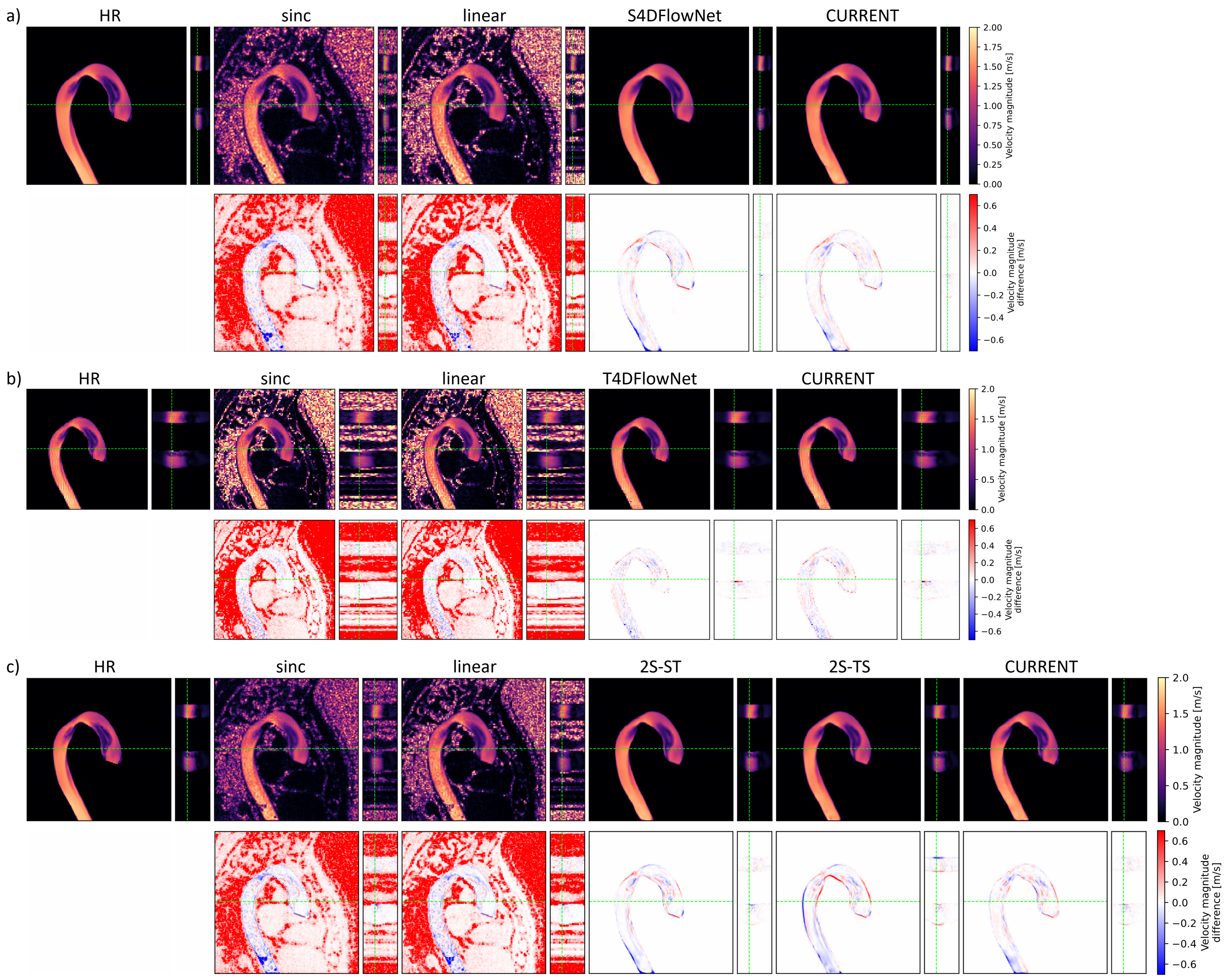

***Figure 3: Qualitative comparison of classical upsampling (sinc and linear) and DL-based SR methods.*** *Velocity magnitudes during peak systole (top rows) and their differences with HR (bottom rows) are shown for a) spatial b) temporal and c) spatiotemporal SR. The dashed green lines in the sagittal (x-y) and temporal (x-t) slices indicate the spatial column and peak systolic frame used for the x-t and x-y slices, respectively.*

***Table 1: Quantitative comparison of spatial SR results.*** *Classical (sinc and linear) and DL-based SR methods (S4DFlowNet and CURRENT) are quantitatively compared (mean ± standard deviation), with the best metrics in bold based on the mean.*

| **Method** | $DE$ (↓) | $DE\text{-}NW$ (↓) | $nRMSE_v$ (↓) | $nRMSE_v\text{-}NW$ (↓) | $\varepsilon_{rel}$ (↓) | $SR_\gamma$ (↑) | $SR_{ED}$ (↓) |
|---|---|---|---|---|---|---|---|
| sinc | 0.273 ± 0.074 | 0.425 ± 0.036 | 1.002 ±0.618 | 2.824 ±1.123 | 6.55 ±7.21 | 0.000 ± 0.000 | 1.000 ± 0.000 |
| linear | 0.232 ± 0.070 | 0.404 ± 0.040 | 0.988 ±0.638 | 2.975 ±1.247 | 7.27 ±4.73 | 0.036 ± 0.013 | 0.951 ± 0.014 |
| S4DFlowNet | 0.097 ± 0.053 | 0.266 ± 0.103 | 0.213 ±0.074 | 0.512 ±0.094 | 5.34 ±5.59 | 0.157 ± 0.072 | 0.890 ± 0.054 |
| CURRENT | **0.072 ± 0.037** | **0.227 ± 0.093** | **0.189 ± 0.067** | **0.463 ±0.097** | **3.72 ±4.06** | **0.192 ± 0.066** | **0.857 ± 0.067** |

***Table 2: Quantitative comparison of temporal SR results.*** *Classical (sinc and linear) and DL-based SR methods (T4DFlowNet and CURRENT) are quantitatively compared (mean ± standard deviation), with the best metrics in bold based on the mean.*

| Method | $DE$ (↓) | $DE$-$NW$ (↓) | $nRMSE_v$ (↓) | $nRMSE_v$-$NW$ (↓) | $\varepsilon_{rel}$ (↓) | $SR^t_\gamma$ (↑) | $SR^t_{ED}$ (↓) |
|---|---|---|---|---|---|---|---|
| sinc | 0.264 ± 0.072 | 0.366 ± 0.051 | 1.139 ±0.784 | 2.098 ±1.226 | 3.40 ±3.94 | 0.040 ± 0.006 | 0.973 ± 0.009 |
| linear | 0.260 ± 0.071 | 0.362 ± 0.052 | 1.125 ±0.774 | 2.071 ±1.210 | 4.19 ±4.39 | 0.236 ± 0.016 | 0.949 ± 0.005 |
| T4DFlowNet | 0.064 ± 0.031 | **0.123 ± 0.057** | **0.202 ± 0.065** | **0.307 ±0.095** | 2.20 ±2.47 | **0.363 ± 0.143** | 0.839 ± 0.091 |
| CURRENT | **0.062 ± 0.031** | 0.127 ± 0.055 | 0.204 ±0.063 | 0.315 ±0.093 | **2.03 ±2.42** | 0.327 ± 0.141 | **0.788 ± 0.101** |

***Table 3: Quantitative comparison of spatiotemporal SR results.*** *Classical (sinc and linear) and DL-based SR methods (2S-ST, 2S-TS, and CURRENT) are quantitatively compared (mean ± standard deviation), with the best metrics in bold based on the mean.*

| Method | $DE$ (↓) | $DE$-$NW$ (↓) | $nRMSE_v$ (↓) | $nRMSE_v$-$NW$ (↓) | $\varepsilon_{rel}$ (↓) | $SR_\gamma$ (↑) | $SR_{ED}$ (↓) | $SR^t_\gamma$ (↑) | $SR^t_{ED}$ (↓) |
|---|---|---|---|---|---|---|---|---|---|
| sinc | 0.254 ± 0.073 | 0.418 ± 0.038 | 0.842 ± 0.504 | 2.431 ±0.925 | 11.28 ±6.96 | 0.000 ± 0.000 | 1.000 ± 0.000 | 0.030 ± 0.012 | 0.977 ± 0.013 |
| linear | 0.211 ± 0.067 | 0.393 ± 0.042 | 0.823 ± 0.514 | 2.535 ±1.017 | 11.09 ±6.11 | 0.032 ± 0.013 | 0.956 ± 0.013 | 0.184 ± 0.063 | 0.960 ± 0.003 |
| 2S-ST | 0.109 ± 0.063 | 0.318 ± 0.134 | 0.230 ± 0.070 | 0.524 ±0.094 | 6.48 ±4.89 | 0.182 ± 0.076 | 0.894 ± 0.055 | 0.323 ± 0.144 | 0.796 ± 0.080 |
| 2S-TS | 0.109 ± 0.076 | **0.244 ± 0.105** | 0.250 ± 0.061 | 0.674 ±0.099 | 6.87 ±4.96 | 0.146 ± 0.057 | 0.898 ± 0.064 | **0.349 ± 0.163** | 0.830 ± 0.097 |
| CURRENT | **0.081 ± 0.040** | **0.244 ± 0.096** | **0.217 ± 0.067** | **0.493 ± 0.092** | **6.22 ±5.08** | **0.184 ± 0.075** | **0.842 ± 0.085** | 0.305 ± 0.148 | **0.788 ± 0.100** |

Using CURRENT for spatiotemporal SR as input to relative pressure estimation showed reduced underestimation (slope closer to identity and lower bias) of relative pressure with less noise (higher $R^2$) compared to separate spatial or temporal SR (Fig. 4).

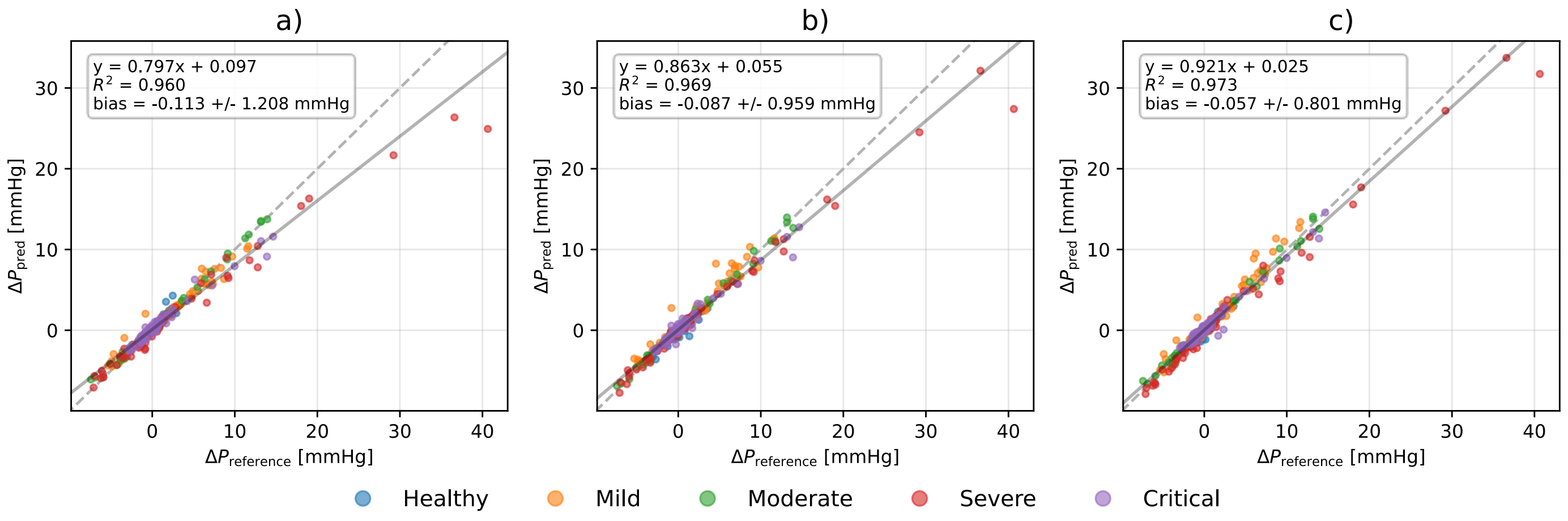

***Figure 4: Linear regression analysis of relative pressure estimation.*** *Relative pressure (ΔP) estimates (between planes in the ascending and descending aorta per cardiac phase) derived from spatiotemporal HR references are compared to ΔP estimates from a) spatial SR with temporal linear interpolation, b) temporal SR with spatial linear interpolation, and c) spatiotemporal SR using CURRENT. Different colors represent different aortic stenosis severities and the dashed and solid black lines represent identity and linear fit (with equation shown in top left textboxes), respectively.*

When using DL-based SR approaches on the in vivo data, reduced velocity noise compared to sinc and linear interpolations was seen (Fig. 5). Note that, in contrast to the synthetic data, ventricular flow was also super-resolved. CURRENT resulted in improved recovery of velocities in the narrow supra-aortic branches compared to S4DFlowNet for spatial SR (Fig. 5a) and compared to both two-stage approaches for spatiotemporal SR (Fig. 5c). Linear regression analysis yielded higher $R^2$ values and correlations closer to identity for CURRENT compared to S4DFlowNet (Fig. 6a). For spatial and spatiotemporal SR, underestimation of high velocities occurred for all SR approaches (Fig. 6a+c) as also shown for peak velocity curves in the ascending aorta (Fig. 7).

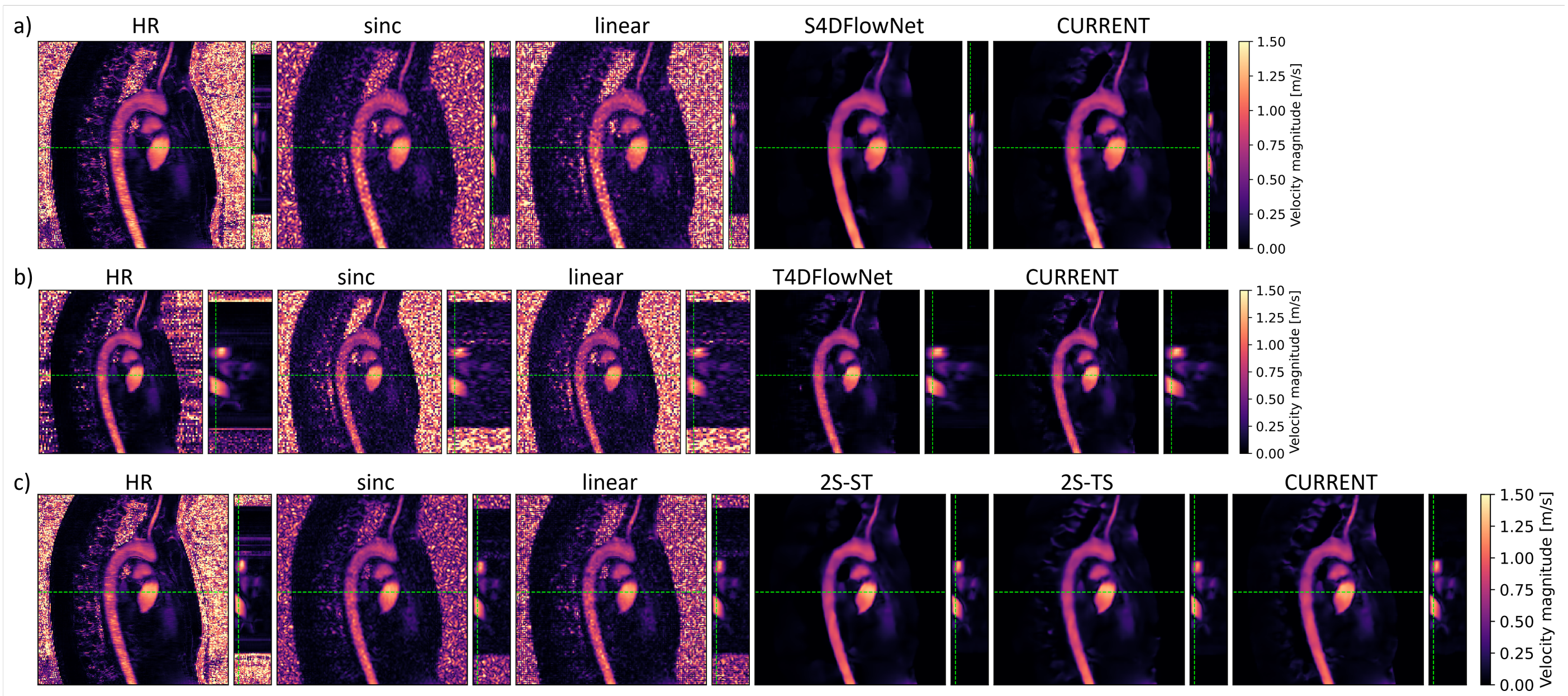

***Figure 5: Qualitative comparison of classical upsampling and SR methods.*** *The velocity magnitudes during peak systole are shown for a) spatial b) temporal and c) spatiotemporal SR. The dashed green lines in the sagittal (x-y) and temporal (x-t) slices indicate the spatial column and peak systolic frame used for the x-t and x-y slices, respectively.*

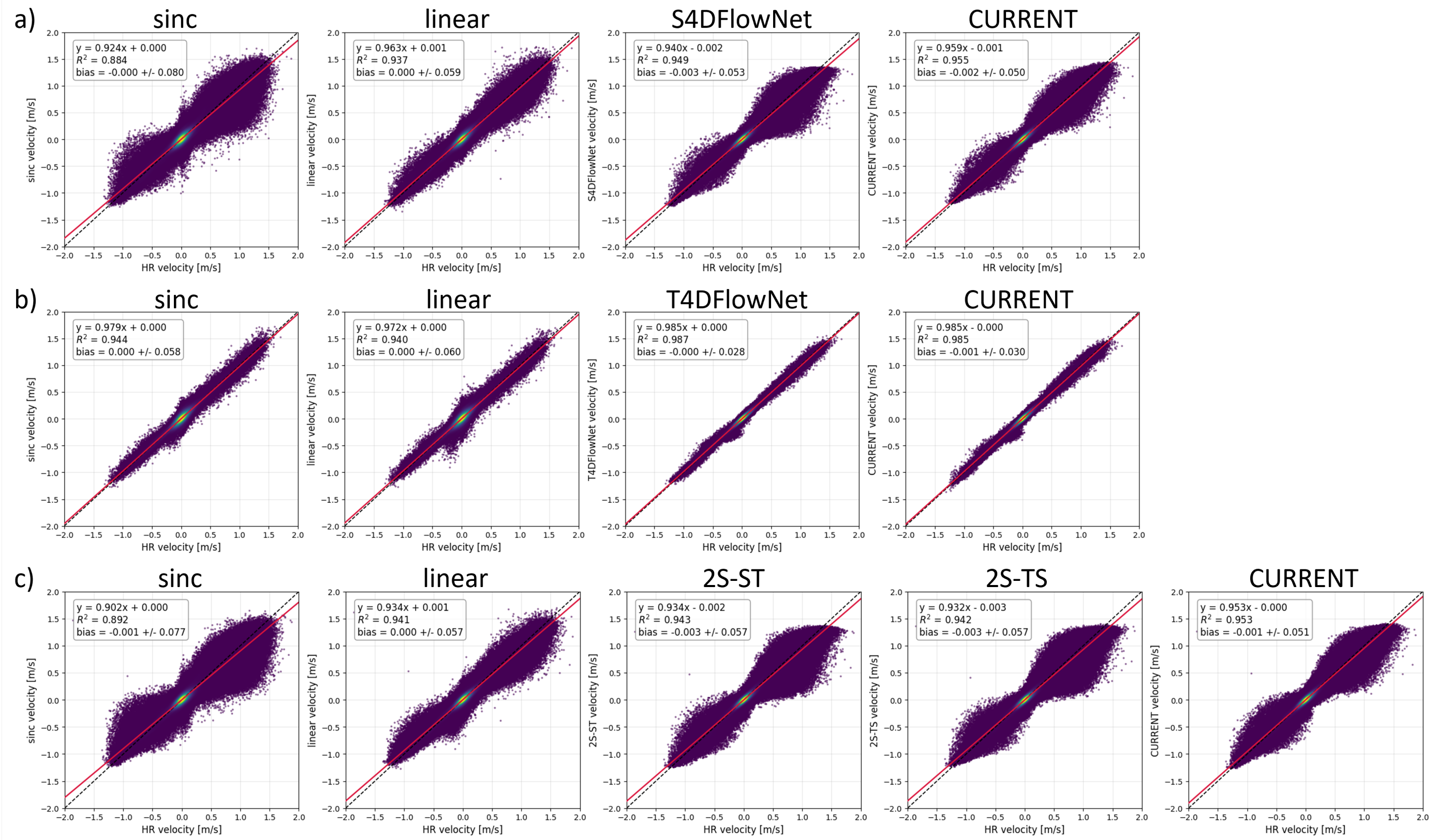

***Figure 6: Linear regression analysis of classical upsampling and SR methods.*** *Velocity components inside the aorta over all cardiac phases are correlated against the HR reference for a) spatial b) temporal and c) spatiotemporal SR. The dashed black and solid red lines represent the identity line and linear fit (with equation shown in top left textboxes), respectively.*

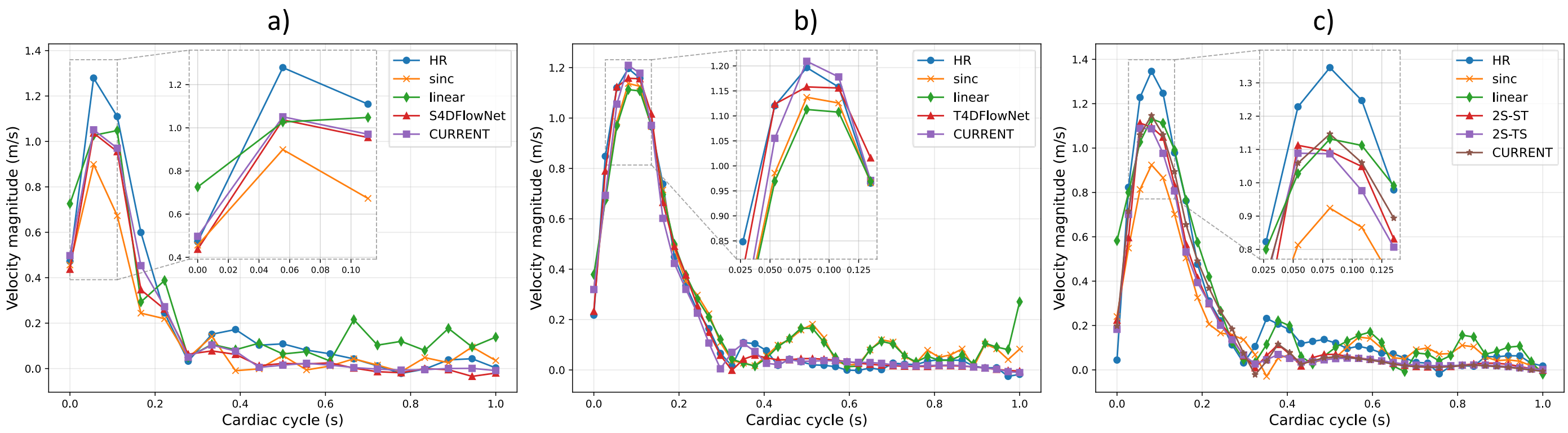

***Figure 7:*** **Comparison of peak velocity curves.** Peak velocity magnitude curves in the ascending aorta are shown for a) spatial b) temporal and c) spatiotemporal SR, with zoomed-in patches of the peak systolic frame.

# 6 Discussion

In this study, a single flexible data-driven recurrent neural network for spatial, temporal, and spatiotemporal SR of 4D flow MRI, named CURRENT, has been proposed. Evaluation on synthetic aortic 4D flow MRI data showed that CURRENT outperforms S4DFlowNet for spatial SR. This improvement is attributed to the exploitation of temporal redundancies across cardiac phases through recurrent connections. For temporal SR, improvements of CURRENT over T4DFlowNet remained negligible. The benefit of spatiotemporal SR became evident for relative pressure estimation, where, compared to either spatial or temporal SR, improved relative pressure maps were obtained. The gains are attributable to improved accuracy of spatial and temporal derivatives as required for the physics-driven estimation using vWERP. Finally, the feasibility of deploying CURRENT on in vivo aortic 4D flow MRI data showed effective denoising of velocity vector fields while maintaining spatiotemporal dynamics.

While the benefit of spatial SR using S4DFlowNet on relative pressure estimation using vWERP has previously been demonstrated for arterial cerebrovasculature [5], the increased pulsatility of aortic flow puts more emphasis on the transient term in the Navier-Stokes equations [38]. Accordingly, temporal velocity gradients become more relevant for relative pressure estimation, demonstrating the need for spatiotemporal SR. In relation to the S4DFlowNet implementation in Ref. [3], which used 71 cardiac frames, our work considered clinically feasible temporal resolutions with 24 cardiac frames. Consequently, increased intra-cardiac-frame motion was present, causing accentuated temporal blurring, which in turn affected the spatial SR task. We also note that the S4DFlowNet implementation used in this work was updated relative to Ref. [3] by not using magnitude data as input and a different loss, as also proposed for T4DFlowNet, as these were found to improve performance.

Compared to tempo-mDCSRN-Flow and TransVort [10], both being LSTMs that exploit the sequential cardiac dimension for improved spatial SR of CSF flow in static cerebral ventricles, our work demonstrated the benefit of RNNs for spatial SR of the thoracic aorta with complex wall motion, where temporal correlations are less trivial, and extended its application to temporal and spatiotemporal SR.

## 6.1 Limitations and future work

While our synthetic data included key features of in vivo 4D flow MRI of the thoracic aorta, some aspects of physical measurements are not yet contained. Among these are MRI system imperfections (such as eddy currents and drifts), the incorporation of actual probe spectra related to particular spatial and velocity encoding gradient waveforms, as well as spatiotemporal jitters which occur due to in vivo heart rate and hemodynamic variation during scanning. To this end, improved domain generalization to in vivo data is to be investigated by considering distributional learning and finetuning of networks on paired in vivo and refined CFD data [39]. Furthermore, the embedding of CFD data into in vivo 4D flow MRI backgrounds resulted in some misalignments between the magnitude and phase values along the vessel wall, which may in turn have led to errors in resolving near-wall velocities. To address this issue, generative data synthesis approaches [40], [41] may be considered, warranting investigation into whether the potential loss of geometric realism could deprecate the generalizability to in vivo inference. Additionally, no physical regularization has been integrated into the network architecture or loss, possibly resulting in flow fields that do not comply with fluid-mechanical principles. Future work could investigate the potential benefits of such regularizations based on e.g. the divergence-free condition and vorticity transport equation [10]. In addition, while SR in 4D flow MRI is currently a post-image reconstruction step, future work is warranted to investigate possible benefits of joint CNN-based reconstruction [31] and SR. Finally, our approach of joint spatiotemporal SR now prompts for comparisons to implicit NN and/or PINN-based SR, which are designed to learn continuous spatiotemporal representations of anatomy [42].

# 7 Conclusions

CURRENT provides a unified recurrent framework for spatial and temporal super-resolution of 4D flow MRI. The proposed approach improves velocity field reconstruction and downstream hemodynamic quantification such as relative pressure estimation, demonstrating the potential of data-driven spatiotemporal super-resolution for accelerated 4D flow MRI.

# 8 Data availability statement

Synthetic thoracic aorta and cerebrovascular 4D flow MRI data with paired high and low resolution data is publicly available at https://gitlab.ethz.ch/ibt-cmr/publications/raclette along with the weights and code for CURRENT at https://gitlab.ethz.ch/ibt-cmr/publications/current.

# 9 Funding

This project was partially supported by grant #2022-274 of the Strategic Focus Area "Personalized Health and Related Technologies (PHRT)" of the ETH Domain (Swiss Federal Institutes of Technology).

# 10 Author contributions

**Luuk Jacobs:** Conceptualization, Data curation, Formal analysis, Investigation, Methodology, Validation, Visualization, Writing – original draft. **Pietro Dirix:** Conceptualization, Data curation, Methodology, Writing – review and editing. **Sebastian Kozerke:** Conceptualization, Funding acquisition, Methodology, Project administration, Resources, Supervision, Writing – review and editing.

# 11 Acknowledgements

We sincerely thank Dr. David Marlevi, Prof. David Nordsletten, Brandon Hardy, and Vincent Lechner for providing and supporting us with the vWERP software.